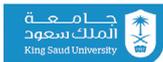

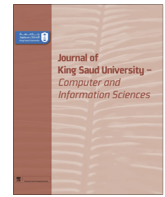

# DQRE-SCnet: A novel hybrid approach for selecting users in Federated Learning with Deep-Q-Reinforcement Learning based on Spectral Clustering

Mohsen Ahmadi [a,*], Ali Taghavirashidizadeh [b], Danial Javaheri [c], Armin Masoumian [d], Saeid Jafarzadeh Ghoushchi [a], Yaghoub Pourasad [e]

[a] Department of Industrial Engineering, Urmia University of Technology (UUT), P.O. Box: 57166-419, Urmia, Iran
[b] Department of Electrical And Electronic Engineering, Islamic Azad University, Central Tehran Branch (IAUCTB), Tehran, Iran
[c] Department of Computer Engineering, Science and Research Branch, Islamic Azad University, Tehran, Iran
[d] Department of Computer Engineering and Mathematics, Universitat Rovira I Virgili, Tarragona, Spain
[e] Department of Electrical Engineering, Urmia University of Technology, Urmia, Iran

## ABSTRACT

Machine learning models based on sensitive data in the real-world promise advances in areas ranging from medical screening to disease outbreaks, agriculture, industry, defense science, and more. In many applications, learning participant communication rounds benefit from collecting their own private data sets, teaching detailed machine learning models on the real data, and sharing the benefits of using these models. Due to existing privacy and security concerns, most people avoid sensitive data sharing for training. Without each user demonstrating their local data to a central server, Federated Learning allows various parties to train a machine learning algorithm on their shared data jointly. This method of collective privacy learning results in the expense of important communication during training. Most large-scale machine learning applications require decentralized learning based on data sets generated on various devices and places. Such datasets represent an essential obstacle to decentralized learning, as their diverse contexts contribute to significant differences in the delivery of data across devices and locations. Researchers have proposed several ways to achieve data privacy in Federated Learning systems. However, there are still challenges with homogeneous local data. This research's approach is to select nodes (users) to share their data in Federated Learning for independent data-based equilibrium to improve accuracy, reduce training time, and increase convergence. Therefore, this research presents a combined Deep-Q-Reinforcement Learning Ensemble based on Spectral Clustering called DQRE-SCnet to choose a subset of devices in each communication round. Based on the results, it has been displayed that it is possible to decrease the number of communication rounds needed in Federated Learning. The realized reduction in the communication rounds are 51%, 25%, and 44% on the three datasets MNIST, Fashion MNIST, and CIFAR-10, respectively.

© 2021 The Authors. Production and hosting by Elsevier B.V. on behalf of King Saud University. This is an open access article under the CC BY-NC-ND license (http://creativecommons.org/licenses/by-nc-nd/4.0/).

## 1. Introduction

With the growth of storage and computational power, data science's relevance has become more evident in engineering systems. Due to the exponential growth of artificial intelligence, machine learning (ML), intelligent manufacturing, and deep learning (DL) in engineering, drastic improvements have been made in recent years (Li et al., 2020a). Mobile devices generate vast amounts of data as computing resources daily. This data can be used in ML systems for various purposes. However, security issues, especially privacy in data transfer to the cloud, deny the possibility of centralized training. In developing data science, however, there

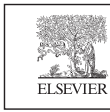

Production and hosting by Elsevier

are two main obstacles. Data governance, which is the most significant factor, is the first obstacle. Through announcing the General Data Protection Regulations, some data will be privatized on the grounds of legal concerns, ensuring that users have full control of their data. No institution or organization has the authority to use the user's data unless it has an agreement.

The second challenge is data accumulation, a confrontational challenge that restricts the growth of the modern business since further data training often increases training efficiency. Besides, in some areas, such as the medical industry, data annotation depends on experienced staff, making accurate data scarce. The absence of labeled data is also a drawback to industrial development. However, there was the advent of Federated Learning (FL) in different parts of engineering systems and business to address these obstacles. FL is a growing ML project that aims to solve the Data Island problem by maintaining data privacy. It refers to various users (such as mobile devices, organizations, institutions) or their nodes synchronized to decentralized ML settings on one or more central servers (McMahan et al., 2018). The main FL processes are shown in Fig. 1. The Federated Averaging (FEDAVG) structure is the foundation of FL. Each computer loads a general global model for local training in the first phase. Second, the global loading model is boosted by multiple local alerts of local data from multiple mobile devices. Then the related slope information is loaded into the cloud in encryption mode. Next, as a new global model, the average upgrade of the local models deployed in the Cloud space will be sent to the system. Lastly, before the optimal output model is obtained or the deadline is hit, the above procedures are replicated.

The implementation of this method would solve the gap between privacy and data sharing for mobile applications. According to its FL properties, the data is not exposed to the third central server; there is a sensitivity to using that data that is a matter of privacy. FL has a lot to do with distributive learning. The traditional distributed system consists of distributed computing with distributed storage operations. FL suggested for design updates for Android clients is somewhat like distributed computing. While FL places excellent focus on privacy, distributed privacy schemes often pay careful attention to recent distributed ML studies. A central processor oversees distributed computing, which connects many computers to multiple locations through a transmission network, with each computer performing different aspects of the same task. FL focuses on creating a collaboration model without privacy leakage, while distributed computing emphasizes on accelerating the processing period. It is necessary to pinpoint the characteristics of FL to distinguish between FL and distributed learning. Data is spread through tens of thousands of mobile computers, or edge nodes, in FL. The total number of nodes in each network is not exceeded by the data. The primary goal of a distributed system is to optimize parallelism in order to minimize the computing or storage load on the central processor. The number of nodes does not exceed FL in the same manner in a distributed system. Today, the world has entered an era of fitness technologies commonly used to track health (Edwards, 2019). Each computer only pro-

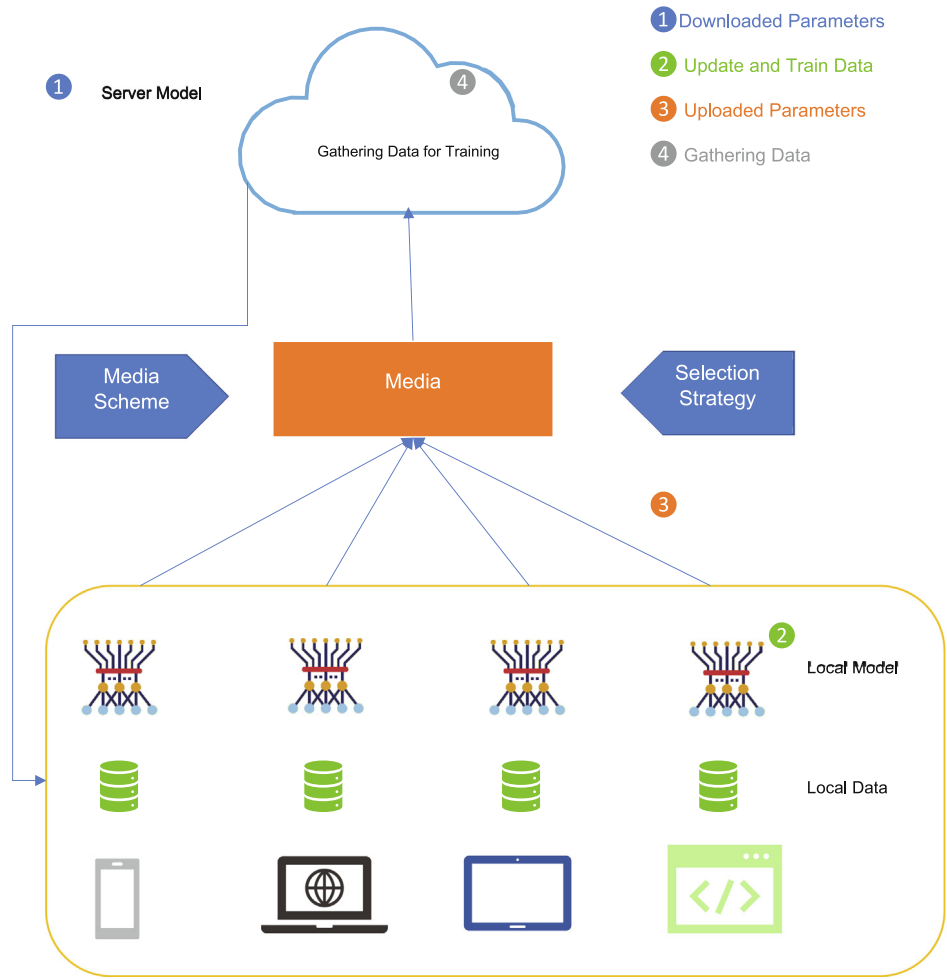

**Fig. 1.** The main processes of FL.

M. Ahmadi, A. Taghavirashidizadeh, D. Javaheri et al.

duces multiple data, and the overall number of devices cannot be compared. FL is more fitting to strengthen the model in this situation.

FL, in comparison to a hierarchical architecture that focuses on the distribution of balanced and uneven data, emphasizes on the distribution of unbalanced and uneven data owing to the variability of device resources. These features include universality for inter-enterprise scenarios, non-IID, decentralized technology, and status equality for each node. This research works in the field of independent wide and unequal distribution. Therefore, it is essential to study their dimensions and angles. FL is described as a newer model of distributed ML systems to use ultimate-user data efficiently. For this training model, all locally trained models are migrated for aggregation to the cloud without users gathering raw data. FL allows a standard model of data vulnerable to privacy to be taught in various applications, including natural language processing, identification of mathematical and visual patterns, artificial vision, and speech recognition. Server-based environments have features such as no wireless connection restrictions, relative stability on mobile devices, and uneven local data distribution. However, distributed ML methods such as FL suffer from these issues, which has created a challenge to describe them (Bonawitz et al., 2019) statistically. ML methods cannot be used simultaneously and in real-time on all mobile systems in a server environment. Therefore, in these methods, it is necessary to select only a few devices at random and be present in the model training cycle, in addition to the overflow of the buffer, to prevent long execution time and instability (Wang et al., 2020). Nevertheless, with all these problems and solutions available, FL still has weaknesses in solving homogeneous local datasets. Moreover, local models that are educated on non-identical independent data can vary considerably. Aggregating these divergent models will dramatically delay the convergence and decrease the model's accuracy substantially. It is necessary to increase FL's efficiency to select stimulus devices or nodes as much as possible. Hence, a DL model is proposed to solve the mentioned challenges that can teach and meet FL's needs. The proposed method selects the best set of devices or nodes in each communication round to balance the amount of bias introduced by independent non-uniform data. It also accelerates and stabilizes the FL process. In the case of extreme paucity of training samples, Salazar et al. suggested a system for resampling the classification model of a predictor. The Network Fourier Transform was employed by the generation block of GAN to create proxies using the vector Markov Random Field theory. The exclusionary phase subsequently uses a linear discriminant on variables to compare the synthetic and original examples' clique similarity (Salazar et al., 2021). For analysing small datasets, Izonin et al. investigated an additional input-doubling technique. The technique for data augmentation within the validation set, both in rows and columns, the use of nonlinear SVR to perform the training procedure, and the formulation of the outcome based on the article's technique were the key phases of the process. The results demonstrate that the suggested data augmentation technique adheres to axial symmetric principles (Izonin et al., 2021).

## 2. Literature review

Various studies in the field of FL have been presented for different purposes. As one of the main researches, which is very similar to the proposed approach of this research (Wang et al., 2020). The optimization and development of FL on non-uniform independent data are performed with the Reinforcement Learning (RL) approach. This study is carried out by evaluating and forming an implied correlation between the distribution of training data on a computer and the weight of the model being trained, indexing the distribution of data based on the loaded model's weight. Applying the Q-Learning technique of RL and further improving it with the DL technique shows that the proposed approach in the three datasets MNIST, Fashion MNIST, and CIFAR, leads to a reduction in the improvement of the use of FL to 49%, 23%, and 42%, respectively. In Rodríguez-Barroso et al. (2020), the FL framework and methodological structure analyzes the software, Sherpa.ai for data privacy based on DL. The compatibility of ML models with FL has been considered an important issue in this research. Moreover, authors in Qi et al. (2020) deals with the use of Block Chains based on FL with privacy to predict traffic flow. FL is intended to solve traffic authentication challenges in predicting traffic flow, and the Block Chains structure seeks to prevent attacks by improving FL's efficiency. Prediction along with improving privacy is the main goal and result of this research. In Dong et al. (2020) an efficient and secure triple FL mechanism called EASTFLY was presented that improves privacy by considering homomorphic cryptography. Improved privacy through FL in the cloud computing environment is also presented in the MNIST data in Fang et al. (2020) with a 20% improvement. FL is also proposed for data processing at the Edge computing level to improve accuracy rates, training time, and improving data aggregation (Jiang et al., 2020). Also, FL scheme with privacy while maintaining integration with a trusted operating environment was presented in Chen et al. (2020) that used neural networks. A multi-layered distributed defense framework to improve instructional data limitations in local nodes using Block Chains and FL is presented in Sharma et al. (2020).

Efficient distributed analysis with FL in Chandiramani et al. (2019) is intended for scalable MNIST data privacy. The placement of authentication-aware services at the Edge computing level has been studied to improve the Quality of Services (QoS) criteria based on FL in Qian et al. (2019). Combining FL with a Greedy algorithm shows improvement in the system. Provides a structure called VerifyNet in Guowen et al. (2019) to improve the security and privacy of data based on FL. Improved authentication and privacy based on FL based on DL is presented in Lyu et al. (2020). Also, in Xiaofeng et al. (2020), FL is used asymmetrically to improve privacy at the level of Edge computing to improve accuracy and reduce local data training time. An approach called AFRL is introduced in Mowla et al. (2020) in the flying ad hoc network (FANET), combining FL and Reinforcement Learning that can greatly neutralize jamming attacks Sattler et al. (2020). FL is used for effective and robust communication that processes non-uniform independent data. The use of MNIST, Fashion MNIST, CIFAR, and KWS data with deep neural networks indicates an improvement in node selection and accuracy. The Sparse Ternary Compression (STC) algorithm is introduced in this paper as a modern compression framework explicitly developed to address the needs of the FL world. The Sparse Ternary Compression (STC) algorithm extends the current high slope scattering compression technique with a new mechanism to allow downstream compression and triple and optimal weight update coding by Golomb. In another similar framework in Zhao et al. (2018), the statistical challenges of FL in independent data are presented independently by presenting a new method. One of the important issues that this research proves is the justification for reducing accuracy with weight divergence, and the simulation results show that accuracy with only 5% of globally shared data can be increased up to 30% for the CIFAR-30 dataset.

In Yeganeh et al. (2020), the inverse distance aggregation for FL is presented with independent non-uniform data that works on medical data at the cloud level. The study of independent non-uniform data structure and methods of applying ML is also presented in Hsieh et al. (2019). This presentation investigates the reduction of computations, improves the accuracy and scalability of its studies, and offers an optimal solution based on DL to solve these problems. FL applications in engineering sciences, especially

M. Ahmadi, A. Taghavirashidizadeh, D. Javaheri et al.

industrial engineering, and its working method have been studied in a review in Li et al. (2020b). The study of security issues and authentication based on FL has been studied in Mothukuri et al. (2021). This research provides a classification and review of approaches and techniques in the field of FL, and by examining security vulnerabilities and threats in FL environments, it tries to identify authentication issues. Other peer-reviewed articles focus on FL, for example, in Yang et al. (2019) the concepts and applications and in Kurupathi and Maass (2020) the study of security and privacy by artificial intelligence methods and many others have been studied. Table 1 indicates a comparison of related methods.

In this paper we presented an approach to select users for sharing the computation in a FL system. The main goal is to improve accuracy, reduce training time, and increase convergence. We presents a hybrid Deep-Q-Reinforcement Learning Ensemble based on Spectral Clustering method to choose a subset of devices in each communication round.

## 3. Proposed method

### 3.1. The theoretical novelty of the research

For the combination of the proposed model with FL, a series of necessary conditions must be considered, including:

- A subset of existing users, each downloading the current model, is chosen.
- Each user in the subset calculates an updated model based on its local data.
- Model updates are sent to the destination from selected clients.
- The server aggregates these models to create an improved global model.

The data such as MNIST, Fashion MNIST, and IFAR-10 among the data of this study are non-IDD. The main difference between this approach and the method presented in Wang et al. (2020) is using a Deep-Reinforcement neural network structure that works in a FL framework, while the paper (Wang et al., 2020) is an RL structure based on FL.

### 3.2. Primary model

DL is ML form that models data patterns into complicated, multi-layered networks and is a more robust and complete distribution than artificial neural networks. Because DL is a very general way of modeling a task, it can solve problems like computer vision and natural language processing that are outside the scope of traditional programs and ML methods. If other techniques fail, DL can yield useful results. It can also build a more reliable model than other methods and minimize the time it takes to produce a reliable model. However, DL models require several computer skills. Another downside of DL is the challenge of decoding DL models. DL's primary feature is that the input and output of the model are over one hidden layer. In most discussions, DL includes the use of deep neural networks. However, many algorithms incorporate DL with several hidden layers rather than neural networks. In essence, other classical ML algorithms run much more rapidly than DL algorithms. Sometimes, a classic model requires one or more CPUs to teach. DL models typically need training and scale stability hardware accelerators like GPUs, TPUs, or FPGAs, and without them, model training would take months. "For several problems, certain classic ML algorithms create a "good enough" model. Other issues have not been solved well by classical ML algorithms in the past. There are several examples of various problems involving DL for generating the right models. Natural language processing is a good thing. For a real problem, a deep neural network can have ten hidden layers. It can be easy or challenging to align. The additional network layers, the more features can be detected. However, the greater the number of network layers, the longer it takes and the harder it becomes to train. In computer vision, the Convolutional Neural Network (CNN) is also used.

**Table 1**
Comparison of the related methods.

| | Year | Method | Purpose and Application |
|---|---|---|---|
| Wang et al. (Wang et al., 2020) | 2020 | Optimization and development of FL in independent non-uniform data with Q-Learning Reinforcement Learning approach | Improve the use of FL in MNIST, Fashion MNIST, and CIFAR-10 data and improve accuracy rates and reduce training time |
| Barroso et al. (Rodríguez-Barroso et al., 2020) | 2020 | Software analysis, Sherpa FL framework, and methodological structures | The adaptability of ML models with FL |
| Qi et al. (Qi et al., 2020) | 2020 | Use Block Chains based on FL with privacy | Predict traffic flow and prevent attacks |
| Dong, e al. (Dong et al., 2020) | 2020 | An efficient and secure triple FL mechanism called EASTFLY based on homomorphic cryptography | Improve privacy of homogeneous local data |
| Fang et al. (Fang et al., 2020) | 2020 | Improve privacy through FL in the Cloud computing environment on MNIST, UCI, and Human Activity Recognition data | Reduce runtime by up to 20% and improve encrypted text transmission by up to 85% accuracy |
| Jiang et al. (Jiang et al., 2020) | 2020 | FL for data processing at the Edge computing level | Improve accuracy rate, training time and improve data aggregation |
| Chen et al. (Chen et al., 2020) | 2020 | Neural Network Based FL Scheme | Privacy by maintaining integrity with a trusted operating environment |
| Sharma et al. (Sharma et al., 2020) | 2020 | Provide a multi-layered distributed defense framework using Block Chains and FL | Improved training data constraints at local nodes |
| Chandiramani, et al. (Chandiramani et al., 2019) | 2019 | Distributed Efficient Analysis with FL | MNIST data privacy on a scalable basis |
| Qian et al. (Qian et al., 2019) | 2019 | Deployment of authentication and privacy-aware services at the Edge computing level with FL based on Greedy algorithm | Improving Quality of Service (QoS) criteria |
| Xu et al. (Guowen et al., 2019) | 2019 | Provide VerifyNet structure based on FL | Improve data security and privacy |
| Lyu et al. (Lyu et al., 2020) | 2020 | FL based on DL | Improve authentication and privacy |
| Lu et al. (Xiaofeng et al., 2020) | 2020 | FL asymmetrically at the level of Edge computing | Improve the privacy and improve accuracy and reduce local data training time |
| Mowla et al. (Mowla et al., 2020) | 2020 | AFRL approach in the flying ad hoc network (FANET) or FANET based on FL and Reinforcement Learning | Detect and neutralize jamming attacks |
| Sattler et al. (Sattler et al., 2020) | 2020 | FL based on deep neural networks for processing non-uniform independent data with sparse ternary compression (STC) algorithm | Practical and robust communication of MNIST, Fashion MNIST, CIFAR, and KWS data - downstream compression capability as well as a triple and optimal Golomb encoding with weight updates |
| Zhao et al. (Zhao et al., 2018) | 2018 | Statistical challenges of FL in independent non-uniform data | Justification of accuracy reduction with weight divergence in the CIFAR-10 data set |
| Yeganeh et al. (Yeganeh et al., 2020) | 2020 | Reverse distance aggregation for FL with non-uniform independent data at the Cloud level | Improve fast processing of medical data |

Natural language and other pattern analyses, such as Long Short-Term Memory (LSTM) and neural attention networks, use the Recurrent Neural Network (RNN). Random forests, also known as random decision forests, are used to solve a wide variety of classification and regression problems. They are not neural networks. They can be used in a lot of deep structures. To model the visual cortex, CNN typically uses innovative, pooling, ReLU, fully connected, and lacking layers. The Convolutional layer essentially integrates various small overlapping areas, and the pooling layer performs a nonlinear sample. The $F(x) = max(0, x)$ the function of activation for ReLU layers is unsaturated. All the activation of the previous layer are linked to fully connected layer. Implementing a SoftMax loss or cross-entropy classification function or a Euclidean loss regression function, a loss layer tests how network training fines the variance between expected and actual tags. Information is converted from input to output in neural feedforward networks through the hidden layers. It limits the interaction between the network and a single state at the same time. The RNN rotates data through a loop which allows for the recall of previous outputs by the network, allowing sequences and time series to be examined. Two typical issues with RNN exist: slopes explode, and the paths disappear (easily removed by closing slopes) (they are not easy to fix). In LSTMs, the network can forget (gateway) or recall previous information by adjusting the weight in both situations. It addresses missing gradients and generates stable LSTM memory for both the long and short term. LSTMs may respond to a long list of previous inputs. Control modules are generalized gates that give the input vector more weight. To process tens of thousands of previous inputs, the hierarchical neural attention encoder employs many layers of attention modules, and to process tens of thousands of previous inputs, the hierarchical neural attention encoder employs many layers of attention modules.

In general, the portion of DRL discussed in this study focuses on Spectral Clustering and has its general mechanism as described in Fig. 2. Where the matrix $M$ is an $n \times d$ dimension. Based on Fig. 2, the following sections show the detailed modeling for this diagram. The proposed approach of this research is to use a DRL model to improve the FL workflow. DL- Reinforcement is a type of RL model that uses DL models to approximate performance for policy functions or values used in RL. The Markov Decision Process (MDP) principle is the foundation for improving learning RL. The MDP consists of a community of countries S, AT, and R. T is a mapping that determines the probability of going to each potential new mode from each action mode pair, and R is a compensation function that correlates each action mode pair with a real value (reward). The T function fits the Markov property in MDP, meaning that the likelihood for change to a novel condition relies on the existing condition and function alone that is not connected to the past. Once the MDP is described, the MDP policy is to map operations from one state to another. The MDP aims to learn the right policy to achieve the best possible reward for each situation. This optimization criterion can be chosen as a single number of advantages, an average level, or a discount level (When advantages are deemed essential than future benefits at present). MDP is a theoretical structure employed to describe the interface variable with an environment where the environment performs both the T and R functions and the policy agent in a series of policy frameworks. The relationship between the agent and the environment is typically broken down by a series of "time steps," in which the agent provides an environmental activity that generates a new status transition and a potential future incentive. Usually, by defining a value function that approximates the value associated with each state, the relation between the optimal criterion and the policy is defined. It would continue according to current policies to estimate the degree of well-being in each situation. A function value can be a V function or a Q function. The V function presumes a value for every State, and Q for each pair of State functions measures a value. The value of the Q function is the sum of the functional status pair plus the value of the V function for the next setting condition (Diro et al., 2017).

Model-based solution models is being applied to find the best approach before the T and R functions in the MDP are well known. These methods are centered on a previously established value function, and certain theoretical results, such as Bellman optimization equations and generalized policy iteration (GPI), are included. In the form of complicated programming solutions: regulation and meaning iteration, these approaches are generally considered two alternative implementations. In the way the GPI system is applied, the distinction between them is. Furthermore, if the above implementation and implementation options for model-driven methods are unavailable, all RL strategies are based on them (model-based or non-model-based). The T and R functions are not specified; non-model-based cleaning techniques should be applied to obtain the optimal strategy. In this case, there are two alternatives: first, after deciding the T and R, you can attempt to learn the T and R models and use the previous methods, and another solution is to specifically try to learn the right policy without knowing the initial meaning of R and T. The current scenario is the case for most RL models, and this is the scenario that has been devised for this purpose. The problem with this scenario is that the lack of knowledge of the R and T dynamics of the environment causes the Bellman equation not to be used, which involves recognizing the probability of moving each state to other potential states. There are several solutions to this: 1) algorithms for temporal difference(TD) learning, 2) strategy gradient, and 3) Monte Carlo methods. Instead of considering all the potential transformations needed by the original Bellman equation, in the Bellman equation, The TD approaches that assume a single state change (from the existing state to the next state under the current policy) have been marginally modified (Diro et al., 2017).

For desirable policy achievement, it is crucial to explore the space of action as much as possible. The main approaches of Greedy exploration are based on the possibility of action. The better action with a probability of p or a random action with a probability (p-1) is chosen for Greedy exploration. Probability-based exploration implies that, concerning this probability distribution, the strategy (directly or indirectly) accounts for any of the proba-bilistic steps that make the sampling process feasible. It can be inferred that the efficiency of learning or the quantity of performance is dependent on various kinds of iterative processes, where a performance state is generated with each iteration. There are two main options for performing these functions: search tables and function approximations. The search table uses a data structure to contain the performance value for every probable mixture of input values (mode and/or output). An output approximation, meanwhile, determines a value for each potential mixture. When the mode and/or operation space is high and has a large effect on efficiency and storage capacity, the choice chosen is significant. Both DRL approaches use feature approximations focused on multiple forms of neural networks (Diro et al., 2017).

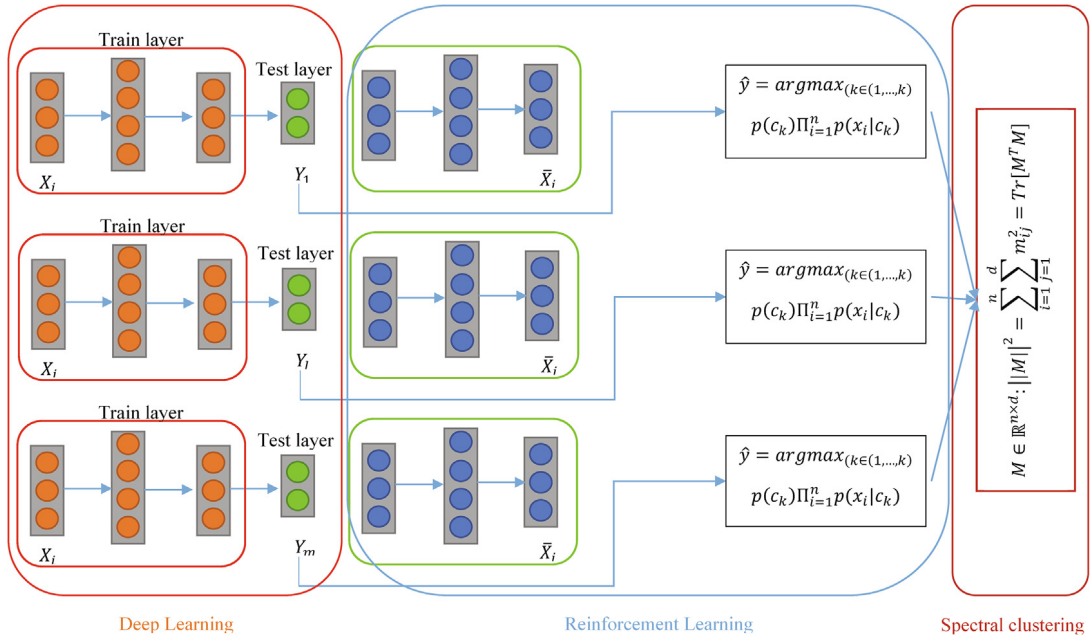

**Fig. 2.** The general Deep-Reinforcement Learning mechanism based on Spectral Clustering presented in this research.

### 3.3. Proposed model of DRL

The DRL hybrid model for improving FL structure is very similar to the method presented in Diro et al. (2017). The only difference is that two neural networks are used in this research model: The existing Q function is executed, while the other targets the Q function. The function Q is intended to replicate the present function Q, which is Q-Learning from the RL family of algorithms but works with delayed coordination due to its presence and combination with a deep neural network. That is, after a certain number of training repetitions, a copy is made. To measure the value of Q for the next case ($\hat{q}_t + 1$), the objective Q function is employed. The purpose of this Q function is to prevent the effect of the moving target when performing a slope of more than $(\hat{q}_t - q_{ref})^2$ and to prevent the return of $q_{ref}$ dependence on the training network. The policy slope is based on a policy performance tutorial called $Q_{current}$. Moreover, using these two determines the operation that must be performed for each possible case. Policy operation is conducted on all layers with a primary multi-layer neural network and activation of ReLU except the last layer that has a SoftMax activation that is a potential distribution of actions or $\pi(a)$. The algorithm begins by predicting actions using policy modes and functions. For all states of a path $s_t$, the action prediction is carried out, and the sequence of anticipated actions is generated. By sampling the likelihood distribution of measures $\pi(a_{\{t\}})$ given by the policy results, these expected measures are retrieved. This section is considered the probability of sample distribution. This analysis uses the $\{T\}$ symbol to denote a sequence in a path's time stages. A series of dimensions, such as $r_{\{T\}}$ or a sequence of vectors, such as $\pi(a_{\{T\}})$ or $\hat{a}_{\{T\}}$ can be used while this symbol is used. Since $\pi(a)$ is the probability vector in the second case for each potential action based on current policy, and $\hat{a}_t$ is an encrypted vector, one of which is assigned to the selected part. Therefore, extending them into a sequence produces vectors. The reward function creates a reward of 0.1; however, in this case, it is a complete set of $\hat{a}_{\{T\}}$ expected actions and $a_{\{T\}}^*$ Ground Truth actions are implemented in one direction. The resulting bonus sequence $r_{\{T\}}$ is translated to the $R_{\{T\}}$ vector of bonus amounts discounted. $R_{\{T\}}$ is calculated with Eq. (1) (Diro et al., 2017).

$$R_{\{T\}} = \left( \sum_{i=0}^{T} \lambda^i r_{t+i}, \ldots, \sum_{i=T}^{T} \lambda^i r_{t+i} \right)$$
$$= \left( \sum_{i=0}^{T} \lambda^i r_{t+i}, \ \sum_{i=1}^{T} \lambda^i r_{t+i}, \ldots, \lambda^T r_{t+T} \right) \qquad (1)$$

It means that each term $R_{\{T\}}$ corresponds to the decreasing amount of consecutive discount bonuses. From the vector of the discounted bonuses $R_{\{T\}}$, the average of the discounted bonuses in different directions $b_{\{T\}}$ is subdivided, and as a result, the vectors of the superiority $A_{\{T\}}$ is obtained. The baseline is also called the vector $b_{\{T\}}$. Advantage values approximate how much better it is for a particular element of the path $s_t$ than the expected return. It is the explanation for subtracting from $R_{\{T\}}$ baseline. The scalar product between the $\pi(a_{\{T\}})$ and $\hat{a}_{\{T\}}$ vector sequences determine the likelihood of a selective action for each step of time $\pi(\hat{a}_{\{t\}})$, because $\hat{a}_t$ is a vector encrypted. A kind of log-loss function is a loss used to train the neural network, which is an estimation of the policy function, with the sum of the reported action likelihood paths performed for a particular path element $\log \pi([\hat{a}_{\{T\}}]_i)$ multiply by the corresponding advantage value $[A_{\{T\}}]_i$. When the training is over, the neural network that executes the policy function is used to predict. For a particular case, the functionality of the distribution policy provides likelihoods for performances. Only the action with the highest probability (no sampling) is selected in this prediction mode. The output of each previous layer and its bias is calculated by the nonlinear actuator function f to create a weighted input $W_n$ for the next layer $n$ of a neural network. In Eq. (2), the loss function is the first part in identifying errors and intrusions using the mean squares error for one k of training data, and the second part is to avoid over-fitting during training (Diro et al., 2017).

$$J(W, b) = \frac{1}{2k} \sum_{i=0}^{k} ||x^{(i)} - \acute{x}^{(i)}||^2 + \frac{\lambda}{2} \sum_{l=1}^{nl-1} \sum_{i=1}^{sl} \sum_{j=1}^{sl+1} \left(W_{ij}^{(l)}\right)^2 \tag{2}$$

In this respect, the number of layers in the deep neural network is represented by $nl$, and the number of neurons in each input layer is represented by $sl$. By combining Eq. (1) and (2), a structure is presented as a combination of Deep-Reinforcement neural networks, the general relation of which will be (3) (Diro et al., 2017).

$$R_{(T)}J(W, b) = ||x^{(i)} - \acute{x}^{(i)}||^2 \cdot \sum_{i=1}^{T} \lambda^i r_{t+i} + \frac{\lambda}{2} \cdot \left(W_{ij}^{(l)}\right)^2 \tag{3}$$

Now the proposed structure separating the work from the model (Diro et al., 2017), is the final modeling of the Deep-Reinforcement model, which should be in the form of an ensemble. In general, the Deep-Reinforcement combined method will be based on the Ensemble structure called DQRE, which can be seen in Eq. (4), its main formula.

$$\acute{y} = argmax_{k \in \{1,...k\}} \ p(c_k) \Pi_{i=1}^{n} p(x_i \mid c_k) \tag{4}$$

In this regard, $p(x)$ means probability $x$. It is data on which the pattern analysis problem should be performed. $p$ is the probability vector for DQRE training operations, $k$ is a factor, and $c$ is a user. Eq. (5) also describes how data is deployed as an operation in the DQRE model.

$$p(C_k|x) = \frac{p(C_k)p(x|C_k)}{p(x)} \tag{5}$$

In Eq. (5), x stands for new data and a vector variable, and C is the class in which data can be placed or not. The probability of an attribute being in a class is equal to the number of similar attributes previously in that class. It is shown in Eq. (6).

$$x = (x_1, \ldots, x_n) \tag{6}$$

Since this system's inputs include a vector of several features, to obtain the probability in the data in the mentioned class, DQRE is used with the adaptive form specified in Eq. (7).

$$\begin{aligned} p(C_k, x_1, \ldots, x_n) &= p(C_k)p(x_1, \ldots, x_n|C_k) \\ &= p(C_k)p(x_1|C_k)p(x_2, \ldots, x_n|C_k, x_1) \\ &= p(C_k)p(x_1|C_k)p(x_2|C_k, x_1)p(x_3, \ldots, x_n|C_k, x_1, x_2) \\ &= p(C_k)p(x_1|C_k)p(x_2|C_k, x_1) \ldots p(x_n|C_k, x_1, x_2, x_3, \ldots, x_{(n-1)}) \end{aligned} \tag{7}$$

Because the probability of attributes occurring in DQRE-trained classes is independent of each other, this independent mode is used for each data as an Eq. (8).

$$\begin{aligned} p(C_k|x_1, \ldots, x_n) &\propto p(C_k, x_1, \ldots, x_n) \\ &\propto p(C_k)p(x_1|C_k)p(x_2|C_k)p(x_3|C_k) \ldots \\ &\propto p(C_k)\Pi_{i=1}^{n} p(x_i|C_k) \end{aligned} \tag{8}$$

The probability of placing $x$ data in class $C$ is calculated using Eq. (9), which can also calculate data properties without using other methods.

$$p(C_k|x_1, \ldots, x_n) = \frac{1}{Z} p(C_k) \Pi_{i=1}^{n} p(x_i|C_k) \tag{9}$$

Because $p(x) = Z$ are constant in all comparisons because the goal is to compare a single $x$ data with several classes and then analyze local data patterns, so its value can be assumed a fixed number and did not consider it. In this way, it obtains the probability that $x$ is in any $C$, and finally assigns any $C$ that has a higher probability of $x$ in the same category. This theorem is based on the Pearson analytical-probabilistic method. This action is specified in Eq. (10).

$$Z = p(x) \tag{10}$$

The target network that works with Q and the predictive network works as Q for the deep section. After identifying the input and applying all the above settings, they need to lead to the goal and prediction, which is the goal of the same reward and prediction of the same penalty. So is $r + \gamma max_{a'}Q(s', a'; \theta_i^-) - Q(s, a; \theta_i)$ and the first part is $r + \gamma max_{a'}Q(s', a'; \theta_i^-)$ for purpose or reward and $Q(s, a; \theta_i)$ for prediction or penalty. In general, the structural flowchart is the DRL Method or DRQ, as shown in Fig. 3.

### 3.4. Spectral clustering

In general, it is stated that the selection of users with Spectral Clustering will be made in the proposed approach. Spectral Clustering strategies without data monitoring are common graph-based approaches to clustering. According to the corresponding similarity matrix's global unique vectors, algorithms for Spectral Clustering typically begin from local data encrypted in a weight graph on data and clusters according to the corresponding similarity matrix's global unique vectors. In Spectral Clustering, for each task, an explicit mapping function is learned simultaneously to predict cluster tags by mapping attributes to the cluster tag matrix. At the same time, the learning process can naturally have differential information to improve clustering performance further. A clustering technique is used by Spectral Clustering, where communities of nodes (i.e., data points) linked or directly adjacent to each other are described in a graph. Then the nodes are drawn into a low-dimensional space that can be divided to form clusters quickly. Spectral Clustering uses information about the specific values (spectra) of specific matrices, such as the Affinity Matrix, Degree Matrix, and Laplacian Matrix derived from a graph or data set. The following are the critical steps in building a Spectral Clustering algorithm (Dr and Renukadevi, 2016):

Wu et al. created an adjustable hybrid clustering technique that integrates unsupervised discrete orthogonal least squares discriminant analysis and discrete Spectral Clustering with adjustable neighbor and side information into a single set. Adaptive weight may be effective production using the suggested framework for quadratic weighted optimization to utilize the given unsupervised approaches (Wu et al., 2021). Belda et al. presented an extended partial coefficient of correlation based on a multivariate Gaussian Mixture Model of the data. To predict the generalized partial correlation coefficients from estimations of the Gaussian Mixture Parameter values, both numerical integration techniques were suggested. As a result, it might be used in any non-Gaussian setting in which the Laplacian matrix is learnt from training signals (Belda

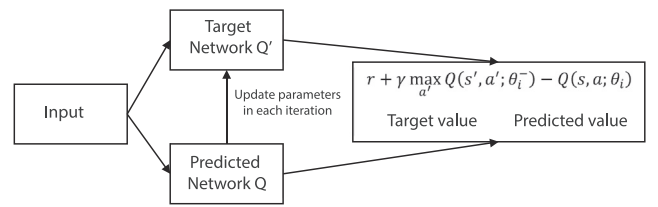

**Fig. 3.** DRQ method flowchart.

et al., 2018). A major difference in user selection in the clustering presented in Wang et al. (2020) and this research is the existence of the clustering problem of the above reference and Spectral Clustering of this research. Therefore, it is necessary to examine the points of difference between these two issues. Spectral Clustering is flexible, allowing non-graphical data to be clustered as well. Spectral Clustering does not make any conclusions about the clusters' shape. Techniques of clustering, such as K-Means, presume that the points allocated to a cluster are around the spherical cluster's center. This is a clear statement that may not be always true. Spectral Clustering allows for building more accurate clusters in such situations. Spectral Clustering can reliably classify cluster observations that also belong to the same cluster, but are further apart due to decreased dimensions than other cluster observations. It should be noted that the Spectral Clustering input of this section is user-focused in one section. Because of this high concentration of users in one section, so that they cannot be placed in different clusters. Therefore, presenting the clustering structure according to Wang et al. (2020) is important. Hence, the Spectral Clustering structure is presented, a deep-reinforcement hybrid model is presented, and by applying a relation, this clustering is determined. This relationship is explained by applying $M \in \mathbb{R}^{n \times d} : ||M||^2 = \sum_{i=1}^{n} \sum_{j=1}^{d} m_{ij}^2 = Tr[M^T M]$ clustering, the different parts of which are presented according to the proposed approach and model. The adjustment of a series of sections in Spectral Clustering in the proposed approach is an important issue that includes the following, which generally changes the clustering problem for users to select from the reference (Wang et al., 2020):

---

**Algorithm I:** The pseudo-code of Spectral Clustering

**Input:** Initial points $S = \{S_1, \ldots, S_2\} \in \mathbb{R}^{n \times d}$
**Output:** Clustering points.

Create $A \in \mathbb{R}$
$D = \sum_j A_{ij}$ as diagonal matrix
$L = D - A$ as Laplacian matrix
$L^{norm} = I - D^{-1/2} A D^{-1/2}$ as normalized Laplacian matrix
Calculate (first $k$ eigenvectors) as $X = \{x_1, \ldots, x_k\} \in \mathbb{R}^{n \times k}$
$Y =$ Normalized $X$ in each row
Consider each row $Y$ as a point in $\mathbb{R}^k$
Cluster points into $k$ clusters using K-Means
Assign principal points $S_i$ to cluster $j$ if column $i$ of the $Y$ matrix is assigned to cluster $j$.

---

**Adjacency and Affinity Matrix (A):** A graph or set of data points can be represented as a neighborhood matrix, where rows and columns show nodes, and the lack or presence of an edge between nodes is demonstrated by inputs. (For example, if the input is in row 0 and column 1 is 1, this indicates that node 0 is connected to node 1). A matrix of MIL is like a matrix of a neighborhood, in that the value of a pair of points shows how close such points are to each other. The combined MIL should be 0 if the pair of points are very different. Also if the points are the same, it could be 1. MIL behaves like the weights at the ends of the graph in this manner.

**Degree Matrix (D):** The Degree Matrix is a diagonal matrix, where the degree of the diagonal node (ie values) is given by the number of edges attached to it. The degree of nodes can also be obtained by considering the sum of each row in the adjacency matrix.

**Laplacian Matrix (L):** The Laplacian Matrix is another representation of graphs or data points related to the aesthetic properties used through Spectral Clustering. One of these representations is obtained by subtracting the neighborhood matrix from the Degree

Matrix (for example $L = D - A$). Eigen values of L are used to gain insight and perform clustering. Some useful items are listed below:

- Spectral Gap: The Spectral Gap is considered the first non-zero eigenvalue. An understanding of the density of the map is given by the spectral gap.
- Fiedler value: The Fiedler value is considered the second eigenvalue and the Fiedler vector is the equivalent vector. Each value in the Fiedler vector gives details as to which side of the decision boundary belongs to a particular node.
- Using L, the first large gap is found between eigenvalues, which generally indicates that the number of eigenvalues before this gap is equal to the number of clusters.

For Spectral Clustering in a FL DQRN Structure For each $M$ matrix, $m_j$ denotes the column vector $j$ and the vector $M, y_i$ stands for the row $Y$ vector, $m_{ij}$ denotes $(i, j)$ the subtraction of the elements $M$ and $Tr[M]$ When tracking, the motion step is $M$, which determines whether this $M$ is a square matrix. $M^T$ represents the displacement matrix $M$. It should be noted that this study uses the $M \in \mathbb{R}^{n \times d} : ||M||^2 = \sum_{i=1}^{n} \sum_{j=1}^{d} m_{ij}^2 = Tr[M^T M]$ matrix norm. It is also assumed that $I$ is suitable as an identifier matrix of the same size. The division of n into k independent clusters at data points $X \in \mathbb{R}^{n \times d}$ is founded on an objective function that reflects the low cluster similarity and the high cluster similarity. The Spectral Clustering algorithm requires the regular Laplace or $L$ graph's eigenvectors in the generic version, which are the relaxations of the index vectors that describe the assignments to a cluster of each data point. It maximizes regular communication in the form of Eq. (11) that also identifies relaxed points.

$$\max_{B \in \mathbb{R}^{n \times k}} Tr(B^T S B) \quad s.t. \quad B^T B = I \tag{11}$$

In this Equation, $S = D^{-1/2} K D^{-1/2} \in \mathbb{R}^{n \times n}$ is the normalized similarity matrix that $K \in \mathbb{R}^{n \times n}$ is the similarity matrix and $D \in \mathbb{R}^{n \times n}$ is the diagonal matrix whose $(i, j)$ subtraction on the elements $X$ is the sum of the i-th column in $X$. Solution for setting the $B \in \mathbb{R}^{n \times n}$ matrix (equal to the eigenvector $k$) corresponds to the largest eigenvalue $k$ of $S$ [75]. After normalizing each column from $B$, the Ensemble Deep Autoencoder Algorithm for each data point $x_i$ from $X$ is assigned to the cluster that column $b_i$ is from $B$, the main difference between this research and the reference [75], which is further, Uses the K-means algorithm. However, this research will use the Ensemble Deep Autoencoder Algorithm. Spectral Clustering works well on custom clusters, unlike many other clustering algorithms, such as the K-means algorithm. Nevertheless, due to the Laplace construction diagram and particular composition's high complexity, this approach's drawbacks are problematic in large-scale handling datasets. In general, by considering and applying a non-independent structure, DRL in Ensemble or DQRE with Spectral Clustering in FL can provide a user selection mechanism during training and testing. More completely, the proposed method is called DQRE-SCnet. In the next section, a simulation of the proposed approach is performed, and the results are discussed in detail; and at the end, a practical comparison with the reference (Wang et al., 2020) is made. In general, the pseudo-code of the proposed approach is as follows:

## 4. Results and discussion

### 4.1. Data collection

An overview of the structure of Non-IID or independent non-uniform data is essential for the final composition. Equal distribu-

M. Ahmadi, A. Taghavirashidizadeh, D. Javaheri et al.

tion means that there is no general trend, and in general, the distribution does not vary significantly, and all the objects in the sample are derived from the same distribution of probability. Independent implies that representations of all independent occurrences are examples of instances. They are, in other words, not related at all. The technical definition of this data is a mathematically uniform distribution of for each $x^i \sim D$ and independent distribution is mathematically $\forall i \neq j, p(x^{(i)}, x^{(j)}) = p(x^{(i)})p(x^{(j)})$. It is essential to sample non-uniform data independent of the training data to ensure that the random slope is an unbiased estimate of the complete slope. Having non-identical independent data on clients means that each subset of data used to update the user locally is statistically identical to the sample drawn and is the same as replacing the entire training set, which is the sum of all local datasets.

### 4.2. Implementation of the findings

Simulation is performed in a MATLAB environment, which can be run after DL packages are installed. The proposed approach uses MNIST, Fashion MNIST, and CIFAR-10 data to ensure performance. A training set of 60,000 cases and an experimental set of 10,000 examples make up the MNIST results. This data is a subset of a larger collection provided by NIST, which includes handwritten characters. The figures have been scaled down and are based on a static picture. The Fashion MNIST data is a series of Zalando article photographs that includes 60,000 training examples and 10,000 research samples. Every example is a 28 × 28 grayscale image with a 10-class mark, each representing data for a single person. The CIFAR-10 dataset includes 60,000 32 × 32-pixel color images divided into ten groups, each with 6,000 images. There are 10,000 sample photos and 5,000 instructional images, each for one user. A set of specific values is used for all three datasets so that the reference (Wang et al., 2020) uses three different values for its approach. The internal structure of DQRE is as shown in Fig. 4.

---

**Algorithm II: The pseudo-code of DQRE-SCnet**

**Input:** Dataset (MNIST, Fashion MNIST, CIFAR-10)
**Output:** Classification network, Best users.

**For**(datasets),
  **For**(user)
    Define (non-iid part for FL);
  **End For**
**End For**
Initialize (agents of DQR)
Define (selection reporting)
X = devices feature;
*Training_set = Separate(X, 0.8)*
*Testing_set = ∀X ≠ xists Traning_set*
**For** (1)
  **DL setting:**
    Define (convolve layer)
    Define (random pooling layer)
    Define (fully connective layer)
  **Reinforcement Q-Learning setting:**
    Define(Q') as target network for reward
    Define(Q) as predicted network
  Ensemble (DL, DQR)
  Solve (DQR)
  Spectral_Clustering(users)
  Label (Rewarded users, Penalty users)
**End For**
Calculate (Evaluation criteria)

---

It should be noted that the structure of the deep-reinforcement neural network based on Spectral Clustering was shown in Fig. 5. According to Fig. 4, in all three data, the DQRE structure uses a torsion layer with windowing 3 × 3 with a descending rate of 24, 18, 12, and 6, and only one random pooling layer is selected by selecting five random attributes. The fully connected layer also has 1 × 1 windowing and rates 7 and 8. In FL, decreasing the number of communication cycles is crucial due to the limited computing capacity and bandwidth of the network and mobile devices. Therefore, the number of communication cycles is used as a measure of performance in DQRE.

The general structure of the proposed approach for selecting users (nodes) in the cloud environment by determining the best accuracy rate and improving training and execution time based on FL based on DRQE-SC is shown in Fig. 6.

After applying the proposed approach and considering the data distribution value $\sigma = 0.5$ as standard in one case, the accuracy is calculated in terms of communication round for the MNIST, Fashion MNIST, and CIFAR-10 datasets as shown in Fig. 6, which is drawn in one place for all data. The reason for drawing them in one place is that they can be compared together, and their distinct separation does not allow a comparison for accuracy in terms of communication round.

In this figure, the number of communication rounds in terms of units is ×100, i.e., 100 communication rounds are considered. Also, on the y-axis, the percentage accuracy is ×100. The number of communication cycles to achieve an accuracy criterion for the proposed approach compared to the reference methods (Wang et al., 2020) is shown in Table 2.

In contrast to centralization learning, which employs possibly the best mini-batches of statistically independent (IID) samples, Federated Learning leverages local information from endpoints, resulting in a wide range of non-IID information. Several of the N users in this scenario seems to have a local data distribution $P_i$ and a local optimal solution $f_i(x) = E(z \sim P_i)[f(x;z)]$, at which $f(x;z)$ is the loss of a model x at an instance z. We usually want to keep $F(x) = 1/N \sum_{i=1}^{N} f_i(x)$ as low as possible. When each $P_i$ is unique, we retrieve the IID configuration. We'll use the letter $F^*$ to represent the lowest value of F found at $x*$. Similarly, we'll use $f_i$ to represent $f^*i$'s minimal value. We consider an occasional communication paradigm, similar to the IID setup, in which $M$ solitary clients engage in each of $T$ rounds, so each client may analyze gradients for K samples throughout each cycle. The change is that at user I the data $z_{i,1}, \ldots, z_{i,K}$ are taken from the client's local service $P_i$. We can't always suppose M = N, unlike in the IID setup, because the user distributions aren't all the same. If a technique relies on M = N, we will remove M and basically put N in the preceding. Though many studies have concentrated on the IID situation, the analytical approach may be generalized to non-IID cases by including a data dissimilarity constraint, such as restricting the variance between customer gradients and the global gradient or the variance between client and global optimal values. The error limitation of local SGD with in non-IID scenario grows worse with this assumption. The amount of local changes K should be less than $T_{1/3}/N$ to meet the pace of $1/\sqrt{TKN}$ (given non-convex goals). Converge rates in non-IID contexts for a (non-complete) collection of federated optimization algorithms. The essential requirements for non-IID contexts, local operations on each user, and other assumptions are summarized in this section. We also show how the method differs from Federated Averaging in terms of convergence rates and the elimination of constants. The complexity of the FEDAVG, K-center, FAVOR and the presented methods are $O(1/T), O(KT), O(k^*/T)$, and $(k^*/T)$ respectively. So that $k^*$ is the number of k selected devices.

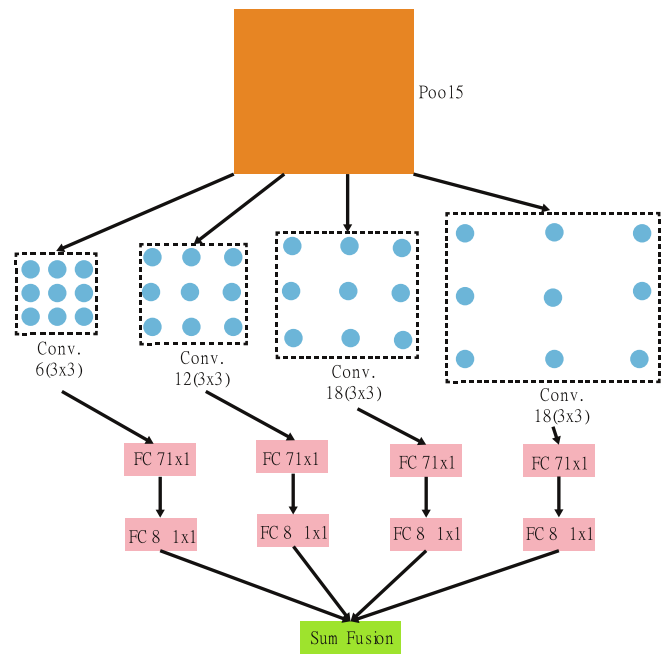

**Fig. 4.** The general structure of DQRE.

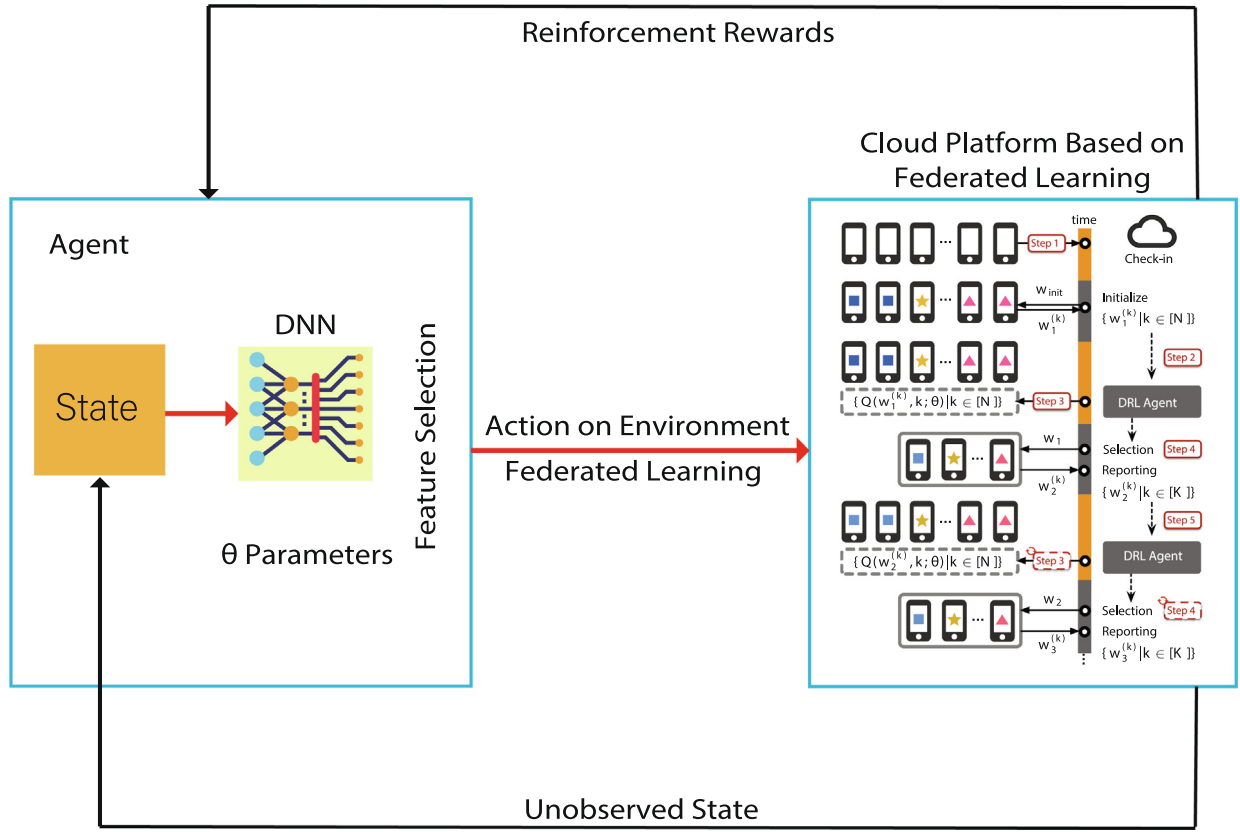

**Fig. 5.** Diagram of the proposed approach (taken from reference (Wang et al., 2020) and FL structures in (McMahan et al., 2018; Edwards, 2019; Bonawitz et al., 2019; Wang et al., 2020; Rodríguez-Barroso et al., 2020; Qi et al., 2020; Dong et al., 2020; Fang et al., 2020; Jiang et al., 2020; Chen et al., 2020; Sharma et al., 2020; Chandiramani et al., 2019; Qian et al., 2019; Guowen et al., 2019; Lyu et al., 2020) and its improvement in this research).

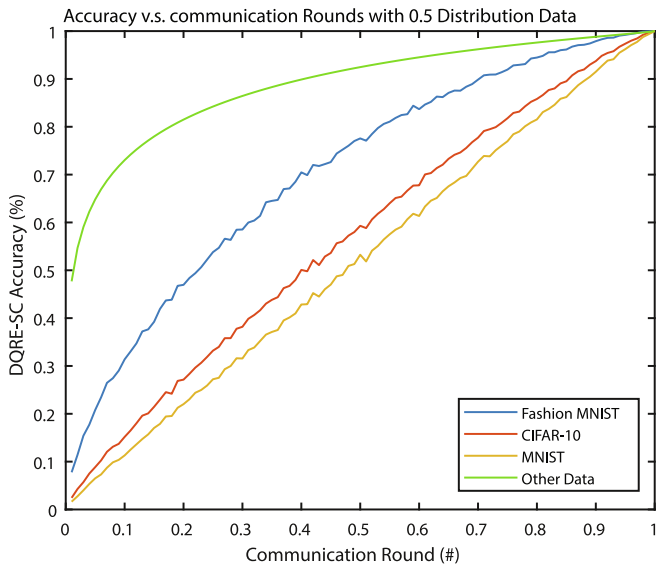

**Fig. 6.** Calculate accuracy in terms of communication round for the proposed approach in three different datasets.

According to the obtained simulations and considering the values of $\sigma$, including 0, H, 0.5, and 0.8, the data were analyzed in general. Based on Table 2, it should be remembered that the numbers in italics show that the formula converges less accurately than the test accuracy. In general, it can be said that DQRE-SCnet on MNIST data with data distribution $\sigma = 1$ equal to 97%, on Fashion MNIST data with data distribution $\sigma = 1$ equal to 88%, and on CIFAR- data 10 with data distribution $\sigma = 1$ is equal to 58%. In general, after applying Spectral Clustering to the DQRE structure, in FL, the resulting spectral clusters may be more biased than the devices or users are chosen by the FEDAVG, K-Center, and FAVOR refer-ences (Wang et al., 2020). Overall, the proposed approach provided 51%, 25, 44% for MNIST, Fashion MNIST, and CIFAR-10 data, respectively, which, compared to the reference (Wang et al., 2020), had 49%, 23%, and 42% results, respectively. It is a better rate for reducing the number of communication rounds. When we train DQRE-SCnet on MNIST data with data distribution $\sigma = 0.8$, the weight of the global model and the weight of the local model is stored in each cycle. The weights of the model stored by the DQRE feature extraction section are reduced to two vectors. By examining the weights of the model trained with Spectral Clustering, it updates the global model by updating the more considerable weight and leading to a greater convergence rate than the methods in Wang et al. (2020). It also creates parallel processing, which speeds up the proposed approach to reduce its computational cost. After all this operation, the DRQE output structure needs a clustering analysis, the output of which is shown for MNIST data as shown in Fig. 7, for Fashion MNIST data as shown in Fig. 8, and for CIFAR-10 data as shown in Fig. 9.

**Table 2**
Calculate accuracy in terms of communication round

| | $\sigma$ | MNIST | Fashion MNIST | CIFAR-10 |
|---|---|---|---|---|
| **Analytical method** | 0 | 55 | 14 | 47 |
| **FEDAVG** (Wang et al., 2020) | 1 | 1517 | 1811 | 1714 |
| **K-Center** (Wang et al., 2020) | 1 | 1684 | 2132 | 1871 |
| **FAVOR** (Wang et al., 2020) | 1 | 1232 | 1497 | 1383 |
| **DQRE-SC** | 1 | 1230 | 1801 | 1694 |
| **FEDAVG** (Wang et al., 2020) | H | 313 | 1340 | 198 |
| **K-Center** (Wang et al., 2020) | H | 421 | 1593 | 188 |
| **FAVOR** (Wang et al., 2020) | H | 272 | 1134 | 114 |
| **DQRE-SC** | H | 304 | 1301 | 100 |
| **FEDAVG** (Wang et al., 2020) | 0.8 | 221 | 52 | 87 |
| **K-Center** (Wang et al., 2020) | 0.8 | 126 | 62 | 74 |
| **FAVOR** (Wang et al., 2020) | 0.8 | 113 | 43 | 61 |
| **DQRE-SC** | H | 111 | 40 | 59 |
| **FEDAVG** (Wang et al., 2020) | 0.5 | 59 | 19 | 69 |
| **K-Center** (Wang et al., 2020) | 0.5 | 67 | 21 | 52 |
| **FAVOR** (Wang et al., 2020) | 0.5 | 59 | 16 | 50 |
| **DQRE-SC** | 0.5 | 57 | 16 | 49 |

The Spectral Clustering structure in DQRE has created two different clusters for MNIST data, and the positive sign as a cluster has a reward, and the middle line is the regression line and separates the two clusters. Fashion MNIST data has a continuous structure, and two clusters with one subset cluster are obtained by the Spectral Clustering method. The first central cluster is the red part on the right and the second central cluster is the blue part on the left. Similarly, the left red cluster is a subset of the right red cluster, and the right blue cluster is a subset of the left blue cluster.

There are two triangular parts for penalty and the positive part for the DQRE model's reward, in which data clustering is done by Spectral Clustering. Moreover, the white midline completely separates the two clusters. Points that are outside the cluster are known as junk data and the same as error data. Finally, the evaluation is done with the ROC chart and AUC rate. The ROC diagram is the proposed approach for MNIST data as shown in Fig. 10, for Fashion MNIST data as shown in Fig. 11, and for CIFAR-10 data as shown in Fig. 12.

This curve is called one of the most important assessment metrics, an indicator of the efficiency of device classification or clustering. In particular, the ROC curve is a graphical illustration of the degree of sensitivity or true prediction versus false prediction in a binary-classification system in which the separation threshold varies. The ROC curve is often seen by plotting the positives properly predicted against the false positive ones predicted. ROC curve area is a number that measures an output component. The Area Under Curve(AUC) has a value from zero to one. A random prediction is a value of 0.5, and values above 0.7 to 1 indicate excellent prediction, classification, or clustering. It is observed that the numerical AUC rate is below one and close to it, which indicates the improvement of the results in terms of AUC and the drawing of the ROC chart and the fit of the proposed approach. Also, other evaluation criteria are shown in Table 3. It should be noted that this analytical output is specified for non-uniform independent data.

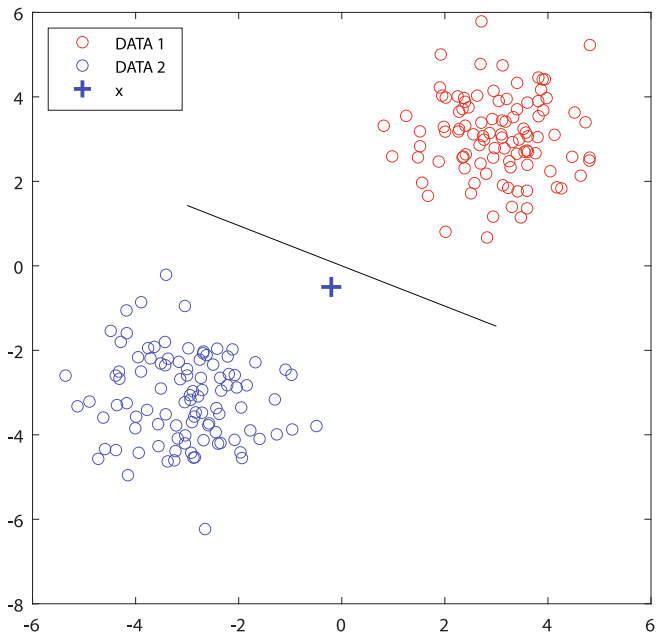

**Fig. 7.** Spectral Clustering of MNIST data in DQRE.

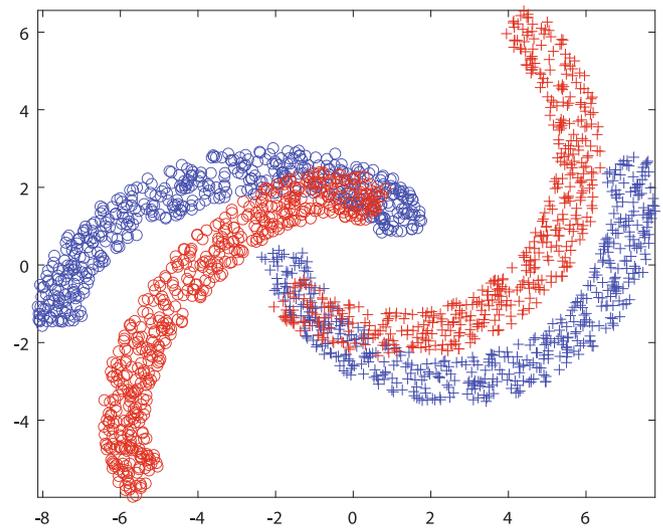

**Fig. 8.** Spectral Clustering of Fashion MNIST data in DQRE.

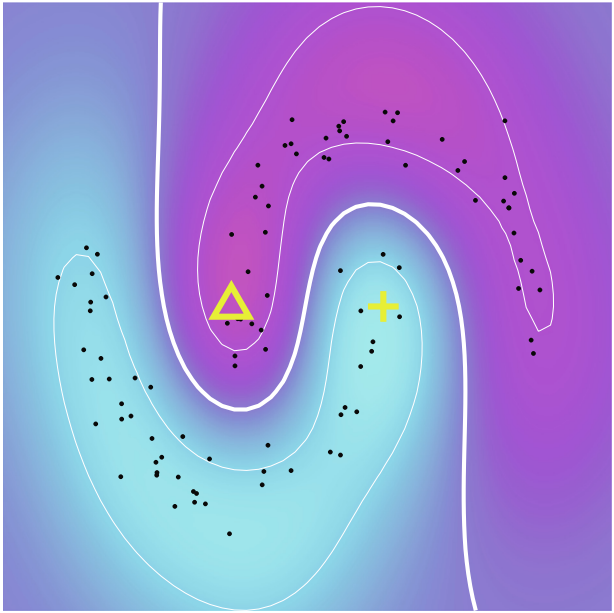

**Fig. 9.** Spectral Clustering of CIFAR-10 data in DQRE.

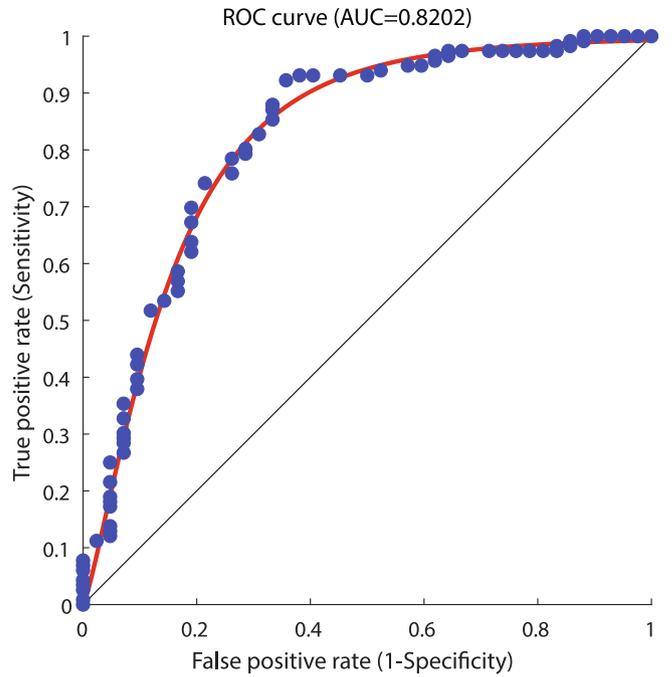

**Fig. 11.** ROC Chart Proposed Approach to Fashion MNIST Data.

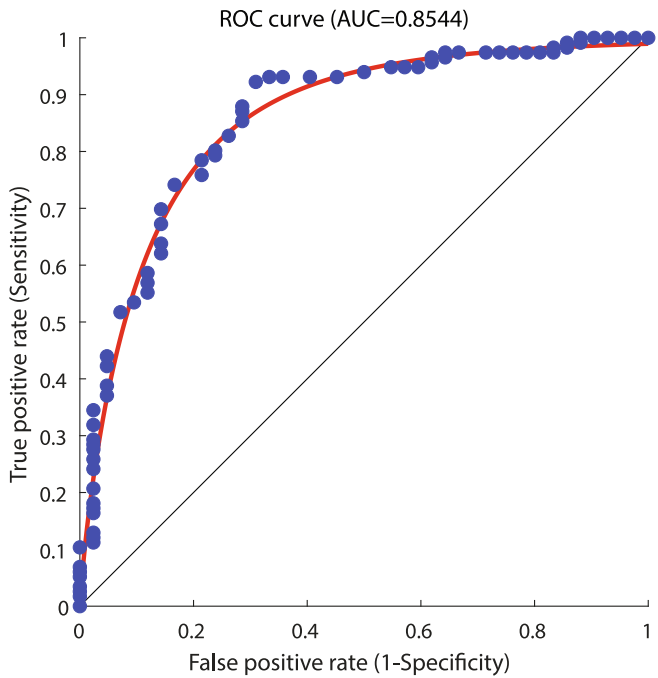

**Fig. 10.** ROC diagram Proposed approach for MNIST data.

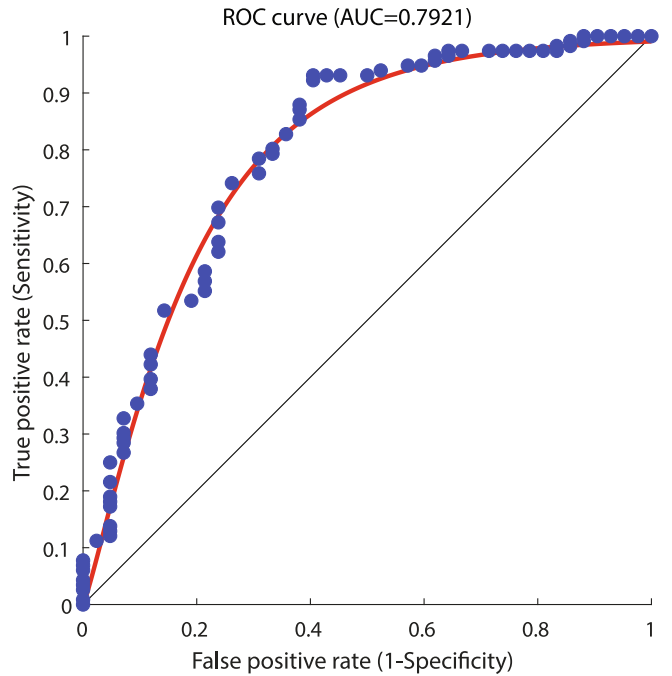

**Fig. 12.** ROC diagram Proposed approach for CIFAR-10 data.

The general results indicate that the proposed approach with data distribution $\sigma = 1$ can achieve good results in terms of evaluation criteria. MNIST and Fashion MNIST datasets have good values, but CIFAR-10 data does not have good accuracy due to high data scatter, and by that, its other evaluation values are very appropriate even if the data distribution is $\sigma = 1$. However, it is still more appropriate and optimal than the reference values (Wang et al., 2020). In addition to offering an effective method for selecting users (customers), this research's ultimate objective is to decrease the number of data training courses. Three MNIST, MNIST Fashion, and CIFAR-10 datasets were considered for this reason. In terms of precision, sensitivity, and other parameters, the proposed approach to the previous methods in Table 3 has achieved significant results that are defined in detail in the previous parts.

**Table 3**
Results of the evaluation criteria of the proposed approach for $\sigma = 1$

| Datasets | Balanced Accuracy | Accuracy | Recall | Kappa | Feature Rate | AUC | Run time (s) |
|---|---|---|---|---|---|---|---|
| MNIST | 94.95 | 97.21 | 92.30 | 97.60 | 91.89 | 0.85 | 65.45 |
| Fashion MNIST | 88.33 | 88.11 | 88.13 | 88.54 | 87.15 | 0.82 | 65.91 |
| CIFAR-10 | 54.20 | 58.42 | 50.01 | 58.38 | 45.60 | 0.79 | 91.56 |

## 5. Future work

The FL implies that any system has an equal chance of participating in and finishing each round. In reality, our framework can implement bias because devices only train when they are connected to a basic network. The models are not used to make user-visible predictions during FL training; instead, once a model is conditioned, it is used to make user-visible predictions. At this stage, if bias in device involvement or other problems result in an inferior model, it will be observed. We have not seen this as a problem in practice yet, but it is undoubtedly application and population-dependent. Future research should focus on quantifying these potential consequences through a broader range of implementations and developing algorithmic or processes methods to minimize them if necessary.

## 6. Conclusion

FL is a modern ML system that helps trustworthy users profit from a mutual learning model's advantages without sharing their private mobile data set explicitly. However, due to insufficient network bandwidth, high communication costs between cloud servers and users are a significant concern. This research begins with selecting the client or nodes or devices while maintaining privacy in a framework to minimize communication rounds and accuracy during FL. For this purpose, homogeneous local data with the independent non-uniform structure are considered, which can increase the number of FL communication rounds by a significant margin. With the help of a mathematical model, an implicit relationship can be found between the model's weight and the data distribution of the trained model. For this purpose, a bias-based structure with the help of independent non-uniform data in each moving device for each client or user (node) with a minimum communication round and improving the accuracy and speed of training in FL is provided. The proposed approach uses the Deep-Reinforcement neural network as an ensemble called DQRE, which can also extract features, which is aimed at the best selection of users. The evaluation uses Spectral Clustering in DQRE, which is considered as DQRE-SCnet. The overall results show that compared to the reference methods (Wang et al., 2020), including FEDAVG, K-Center, and FAVOR in the MNIST data set, Fashion MNIST and CIFAR-10 have better rates. The values obtained by applying the DQRE algorithm in FL for MNIST, Fashion MNIST, and CIFAR-10 data with data distribution $\sigma = 1$ are 97%, 88%, and 58%, respectively, is a relative improvement over Reference (Wang et al., 2020) has been obtained. This improvement can be achieved in the evaluation phase by applying Spectral Clustering to DQRE, which for MNIST, Fashion MNIST, and CIFAR-10 data is 51%, 25, 44%, respectively, compared to the reference (Wang et al., 2020), which results in 49%, 23% and 42%, is relatively more optimal. The presented study involved with several limitations. First, the unavailability real devises for implementing the techniques. Moreover, some problems involved with high computation time process for any datasets. Also, big machine learning networks involved high complexity.

## Declaration of Competing Interest

The authors declare that they have no known competing financial interests or personal relationships that could have appeared to influence the work reported in this paper.

## References

Bela, J., Vergara, L., Salazar, A., Safont, G., 2018. Estimating the Laplacian matrix of Gaussian mixtures for signal processing on graphs. Signal Processing 1 (148), 241–249.

Bonawitz, K., Eichner, H., Grieskamp, W., Huba, D., 2019. Towards Federated Learning at Scale: System Design. In: Proc. the Conference on Systems and Machine Learning (SysML).

Chandramani, Kunal, Garg, Dhruv, Maheswari, N., 2019. Procedia analysis of distributed and federated learning models on private data. Procedia Computer Science 165, 349–355.

Chen, Yu., Luo, Fang, Li, Tong, Xiang, Tao, Liu, Zheli, Li, Jin, 2020. A training-integrity privacy-preserving federated learning scheme with trusted execution environment. Information Sciences 522, 69–79.

Abebe Abeshu Diro, Naveen Chilamkurti, Distributed attack detection scheme using deep learning approach for Internet of Things: Future Generation Computer Systems, 12 2017..

Dong, Ye, Chen, Xiaojun, Shen, Liyan, Wang, Dakui, 2020. EaSTFLy: Efficient and secure ternary federated learning. Computers & Security 94, 101824.

Dr. S. Meenakshi, Renukadevi, R., 2016. A review on spectral clustering and its applications. International Journal of Innovative Research in Computer and Communication Engineering 4 (8), 14840–14845.

Edwards, J., 2019. Medicine on the Move: Wearable devices supply health-care providers with the data and insights necessary to diagnose medical issues and create optimal treatment plans [Special Reports]. IEEE Signal Processing Magazine 36 (6), 8–11.

Fang, Chen, Guo, Yuanbo, Wang, Na., Ankang, Ju., 2020. Highly efficient federated learning with strong privacy preservation in cloud computing. Computers & Security 96.

Guowen, Xu., Li, Hongwei, Liu, Sen, Yang, Kan, Lin, Xiaodong, 2019. VerifyNet: Secure and verifiable federated learning. IEEE Transactions on Information Forensics and Security 15, 911–926.

Kevin Hsieh, Amar Phanishayee, Onur Mutlu, Phillip B. Gibbons, 2019. The Non-IID Data Quagmire of Decentralized Machine Learning. arXiv:1910.00189..

Izonin, I., Tkachenko, R., Shakhovska, N., Lotoshynska, N., 2021. The additive input-doubling method based on the svr with nonlinear kernels: small data approach. Symmetry 13 (4), 612.

Hui Jiang, Min Liu, Bo Yang, Qingxiang Liu, Jizhong Li, Xiaobing Guo, Customized Federated Learning for accelerated edge computing with heterogeneous task targets. Computer Networks 183 107569..

Kurupathi, Sheela Raju, Maass, Wolfgang, 2020. Survey on federated learning towards privacy preserving AI. DFKI LogoDeutsches Forschungszentrum für Künstliche Intelligenz, German Research Center for Artificial Intelligence.

Li, L., Wang, Y., Lin, K., 2020a. Preventive maintenance scheduling optimization based on opportunistic production-maintenance synchronization. Journal of Intelligent Manufacturing.

Li, Li, Fan, Yuxi, Tse, Mike, Lin, Kuo-Yi, 2020b. A review of applications in federated learning. Computers & Industrial Engineering 149, 106854.

Lingjuan Lyu, Jiangshan Yu, Karthik Nandakumar, Yitong Li, Xingjun Ma, Jiong Jin, Han Yu, Kee Siong Ng, 2020. Towards Fair and Privacy-Preserving Federated Deep Models. IEEE Transactions on Parallel and Distributed Systems 31(11) 2524–2541..

McMahan, H.B., Zhang, L., Ramage, D., Talwar, K., 2018. Learning differentially private recurrent language models. ICLR.

Viraaji Mothukuri, Reza M. Parizi, Seyed amin Pouriyeh, Yan Huanga, Ali Dehghantanha, and Gautam Srivastava. A survey on security and privacy of federated learning. Future Generation Computer Systems 115 619–640..

Mowla, Nishat I., Tran, Nguyen H., Doh, Inshil, Chae, Kijoon, 2020. AFRL: Adaptive federated reinforcement learning for intelligent jamming defense in FANET. Journal of Communications and Networks 22 (3), 244–258.

Yongfeng Qian, Long Hu, Jing Chen, Xin Guan, Mohammad Mehedi Hassan, and Abdulhameed Alelaiwi. Privacy-aware service placement for mobile edge computing via federated learning. Information Sciences 505 562–570..

Yuanhang Qi, M. Shamim Hossain, Jiangtian Nie, Xuandi Li, 2020. Privacy-preserving blockchain-based federated learning for traffic flow prediction. Future Generation Computer Systems, Available online Dec. 10, 2020, In Press, Journal Pre-proof..

Nuria Rodríguez-Barroso, Goran Stipcich, Daniel Jiménez-López, José Antonio Ruiz-Millán, Eugenio Martínez-Cámara, Gerardo González-Seco, M. Victoria Luzón, Miguel Angel Veganzones, Francisco Herrera, 2020. Federated Learning and Differential Privacy: Software tools analysis, the Sherpa.ai FL framework and methodological guidelines for preserving data privacy. Information Fusion 64 270–292..

Salazar, A., Vergara, L., Safont, G., 2021. Generative Adversarial Networks and Markov Random Fields for oversampling very small training sets. Expert Systems with Applications 1, (163) 113819.

Sattler, Felix, Wiedemann, Simon, Müller, Klaus-Robert, Samek, Wojciech, 2020. IEEE Transactions on Neural Networks and Learning Systems 31 (9), 3400–3413.

Pradip Kumar Sharma, Jong Hyuk, and Park Kyungeun Cho, 2020. Blockchain and federated learning-based distributed computing defence framework for sustainable society. Sustainable Cities and Society 59..

Hao Wang, Zakhary Kaplan, Di Niu, Baochun Li, 2020. Optimizing Federated Learning on Non-IID Data with Reinforcement Learning. IEEE INFOCOM 2020 – IEEE Conference on Computer Communications, Toronto, ON, Canada, Canada..

Wu, T., Zhang, R., Jiao, Z., Wei, X., Li, X., 2021. Adaptive Spectral Rotation via Joint Cluster and Pairwise Structure. IEEE Transactions on Knowledge and Data Engineering.

Xiaofeng, Lu., Liao, Yuying, Lio, Pietro, Hui, Pan, 2020. Privacy-preserving asynchronous federated learning mechanism for edge network computing. IEEE Access 8, 48970–48981.

Qiang Yang, Yang Liu, Tianjian Chen, and Yongxin Tong. Federated Machine Learning: Concept and Applications. ACM Transactions on Intelligent Systems and Technology, Volume 10, No. 2, 2019..

Yousef Yeganeh, Azade Farshad, Nassir Navab, Shadi Albarqouni, Inverse Distance Aggregation for Federated Learning with Non-IID Data. MICCAI Workshop on Domain Adaptation and Representation Transfer, DART 2020, DCL 2020: Domain Adaptation and Representation Transfer, and Distributed and Collaborative Learning, 2020 150–159..

Yue Zhao, Meng Li, Liangzhen Lai, Naveen Suda, Damon Civin, Vikas Chandra, Federated Learning with Non-IID Data. arXiv:1806.00582, 2018..